\documentclass[a4paper, 10pt, conference]{ieeeconf}

\IEEEoverridecommandlockouts                              
\usepackage{graphics}
\usepackage{epsfig}
\usepackage{mathptmx}
\usepackage{times}
\usepackage{amssymb}  
\usepackage{amsmath,graphicx}
\usepackage{diagbox}
\usepackage{amssymb}
\usepackage[usenames,dvipsnames,svgnames,table]{xcolor}
\usepackage{bm}
\usepackage{tabularx}
\usepackage{cases}
\usepackage{booktabs}
\usepackage{epstopdf}
\usepackage{courier}
\usepackage{extarrows}
\usepackage{enumerate}
\usepackage[mathscr]{eucal}
\title{\LARGE \bf
A Historical Review of Forty Years of Research on CMAC
}

\author{Frank Z. Xing\\
\normalsize{School of Computer Science and Engineering}\\
Nanyang Technological University\\ \texttt{zxing001@e.ntu.edu.sg}\\
}

\begin{document}

\maketitle
\thispagestyle{empty}
\pagestyle{empty}

\begin{abstract}
The Cerebellar Model Articulation Controller (CMAC) is an influential brain-inspired computing model in many relevant fields. Since its inception in the 1970s, the model has been intensively studied and many variants of the prototype, such as KCMAC, MCMAC, and LCMAC, have been proposed. This review article focus on how the CMAC model is gradually developed and refined to meet the demand of fast, adaptive, and robust control. Two perspective, CMAC as a neural network and CMAC as a table look-up technique are presented. Three aspects of the model: the architecture, learning algorithms and applications are discussed. In the end, some potential future research directions on this model are suggested.
\end{abstract}

\vspace{0.3cm}
\section{INTRODUCTION} \label{intro}
The Cerebellar Model Articulation Controller (CMAC) was proposed by J. S. Albus in 1975 \cite{albus75}. Parallel at this time in the history, the concept of perceptron \cite{rosen61} had already been popular, whereas effective learning schemes to tune perceptrons \cite{rume86} were not on the stage yet. In 1969, Minsky and Papert also pointed out the limitations that the exclusive disjunction logic cannot be solved by the perceptron model in their book \textit{Perceptrons: An Introduction to Computational Geometry}. These facts made it less promising to consider CMAC as a neural network form. Consequently, although the name of CMAC appears bio-inspired enough, and the theory that the cerebellum is analogous to a perceptron has been proposed earlier \cite{albus71}, CMAC was emphasized to be understood as a table referring technique that can adaptive to real-time control system. Nevertheless, the underlying bioscience mechanism was addressed again in 1979 by Albus \cite{albus79}, which always gives the CMAC model two different ways of interpretation. 

\begin{figure}[thpb]
\centering
\includegraphics[scale=0.65]{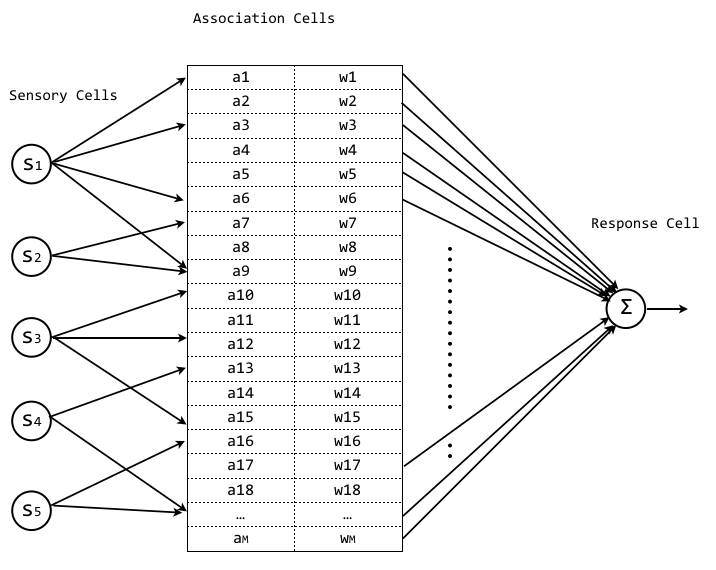}
\caption{The primary CMAC structure}
\label{pcmac}
\end{figure}

The structure of CMAC was originally described as two inter-layer mappings illustrated in Fig. \ref{pcmac}. The control functions are represented in the weighted look-up table, rather than by solution of analytic equations or by analog \cite{albus75}.
If we use $\mathit{S}$ to denote sensory input vectors, and let $\mathit{A}$ and $\mathit{P}$ stand for association cells and response output vectors respectively, both CMAC and multilayer percptrons (MLP) model can be formalized as:
\begin{numcases}{}
f: \mathit{S} \rightarrow \mathit{A} \nonumber \\
g: \mathit{A} \rightarrow \mathit{P}\nonumber
\end{numcases}
the final output can be calculated as:
\[
y = \sum \mathit{A}^*_i w_i
\]
where the asterisk denotes a link or activation between certain association cell and the output. The nuance between MLP and CMAC model is that, for mapping $f$, MLP model is fully connected but CMAC restricts the association in a certain neighboring range. This property of mapping significantly accelerates the learning process of CMAC, which is considered a main advantage of it comparing to other neural network models. 

It is notable that CMAC may not represent accurately how the human cerebellum works, even at a most simplified level. For instance, recent biological evidence from eyelid conditioning experiments suggests that the cerebellum is capable of computing exclusive disjunction \cite{voicu08}. However, CMAC is still an important computational model, because the restriction on mapping function effectively decreased the chance of been trapped in a local minimum during learning process.

Despite the advantage aforementioned, CMAC has the following  disadvantages as well \cite{smith98}:
\begin{itemize}
\item many more weight parameters are needed comparing to normal MLP model
\item local generalization mechanism may cause some training data to be incorrectly interpolated, especially when data is sparse considering the size of CMAC 
\item CMAC is a discrete model, analytical derivatives do not exist
\end{itemize}
As a response to these problems, modified or high order CMAC, storage space compressing, and fast convergent learning algorithms are continuously studied. These discoveries will be elaborated in the following sections. Recent advances in Big Data and computing power seem to have watered down these problems. But in many physical scenarios, the computing power are still restricted and high speed responses are required. This serves for the reason to further study CMAC-like models, though it has been in and out of fashion for several times.

Although there has been few other pioneer review works on CMAC, for instance by Mohajeri et al. in 2009 \cite{moha09}, this article is inventive for its chronological perspective. rather than emphasizing on detailed techniques. The remainder of the article is organized as follows: Section \ref{archi} provides the evolution trajectory of CMAC structure and efficient storage techniques. Section \ref{lrn} discusses the learning algorithms. Section \ref{app} presents various circumstances that CMAC model has been applied. Section \ref{disc} summarizes the paper, and instigates discussions about potential improvements that can be made on CMAC model.

\section{ARCHITECTURE} \label{archi}
\subsection{Basic Architectures}
\begin{figure}[thpb]
\centering
\includegraphics[scale=0.4]{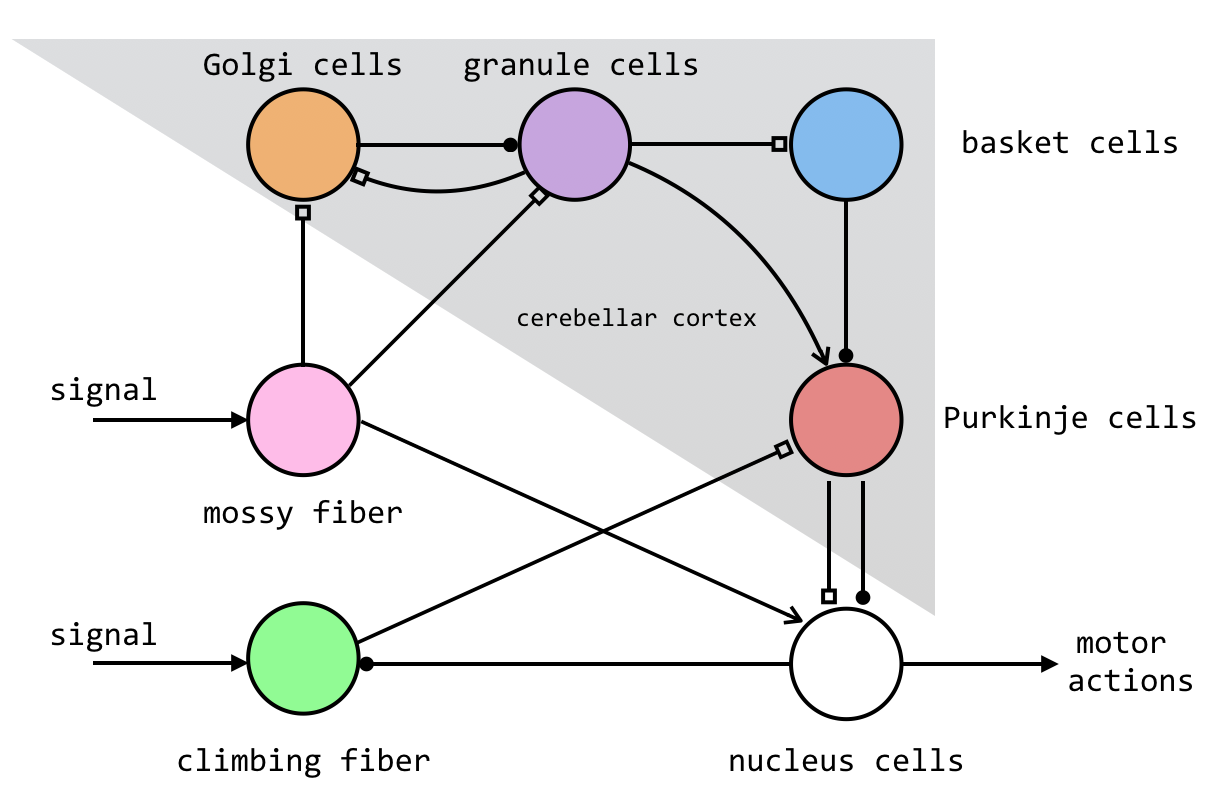}
\caption{Mauk's model assumes complex interaction between different type of cells: open arrow stands for plastic synapses; filled circle arrow stands for inhibitory synapses; hollow square arrow stands for excitatory synapses.}
\label{cerebellum}
\end{figure}

\begin{figure}[thpb]
\centering
\includegraphics[scale=0.35]{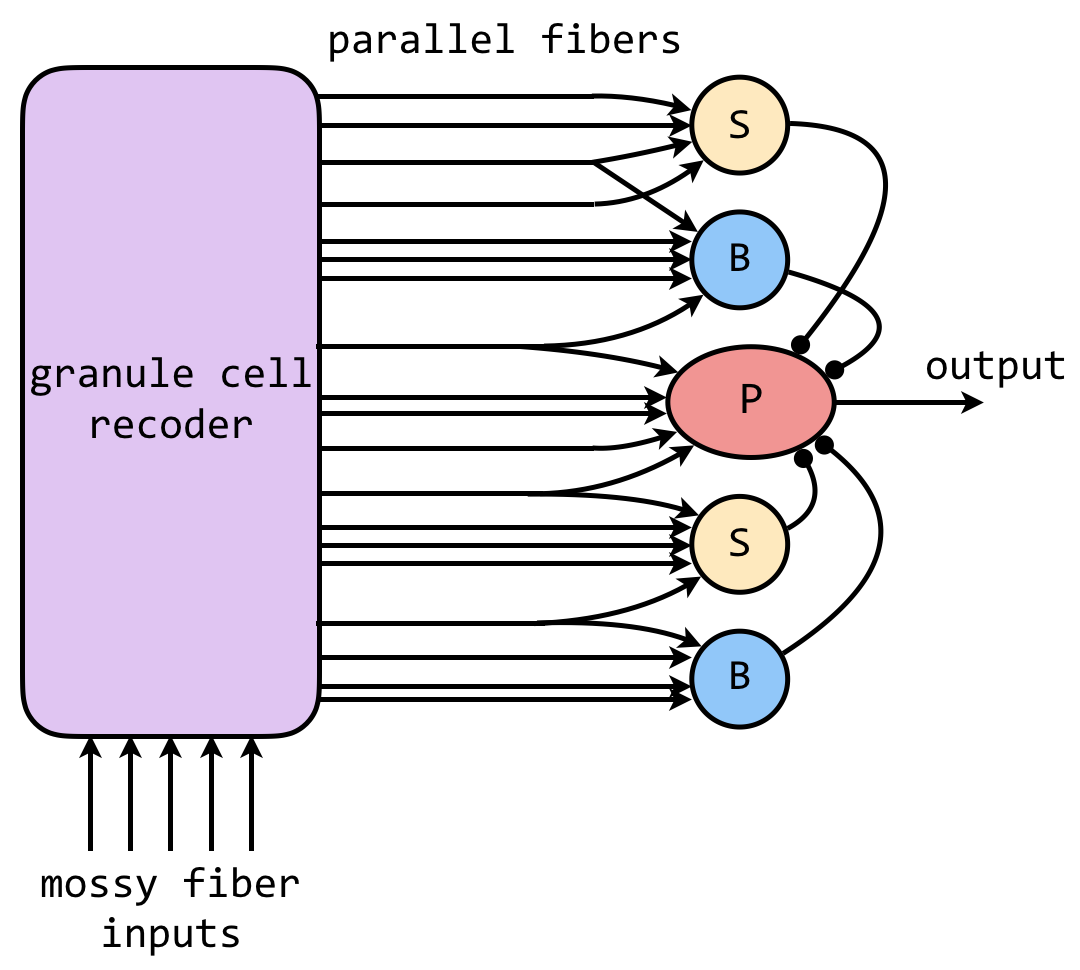}
\caption{Albus' model \cite{albus71} of cerebellum resembles a forward feed perceptron. It assumes climbing fiber activity would cause parallel fiber synapses to be weakened. S stands for stellate b cells, B stands for basket cells}
\label{cerealbus}
\end{figure}

Before the CMAC was proposed in the 1970s, the anatomy and physiology of cerebellum has been studied for a long time. It is widely agreed that many different type of nerve cells are involved in cerebellar functioning. Fig. \ref{cerebellum} shows the computational model proposed by Mauk and Donegan \cite{mauk}. A more simple and implementable model proposed by Albus \cite{albus79} is exhibited in Fig. \ref{cerealbus}.

\begin{figure}[thpb]
\centering
\includegraphics[scale=0.47]{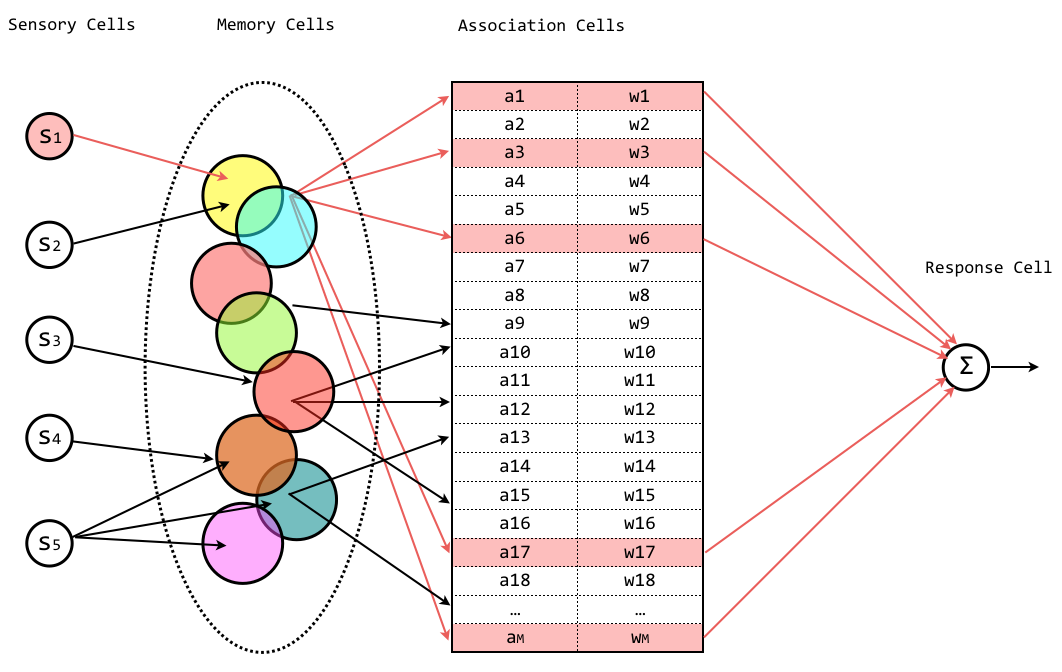}
\caption{CMAC with conceptual memory}
\label{newcmac}
\end{figure}

Based on the computational model of cerebellum (Fig. \ref{cerealbus}), the primary CMAC structure as shown in Fig. \ref{pcmac} is conceived. While it is obvious that high dimensional proximity of association rules cannot be captured in this primary form, because the nodes are arranged into one dimensional array. A simple solution to this problem is to introduce some nonlinearity to the first mapping. As a result, another layer called ``conceptual memory'' was soon added to the CMAC structure, which involves one additional mapping to the primary structure. The function of conceptual memory is illustrated in Fig. \ref{newcmac}.

If we use $\mathit{A}$ to represent the actual memory (Association Cells), $\mathit{M}$ is the conceptual memory to encode $\mathit{S}$. Then the conceptual mapping $f$ is more sparse and constrained within a certain range, but mapping $g$ could be random.
\begin{numcases}{}
f: \mathit{S} \rightarrow \mathit{M} \nonumber \\
g: \mathit{M} \rightarrow \mathit{A}\nonumber\\
h: \mathit{A} \rightarrow \mathit{P}\nonumber
\end{numcases}

\begin{figure}[thpb]
\centering
\includegraphics[scale=0.35]{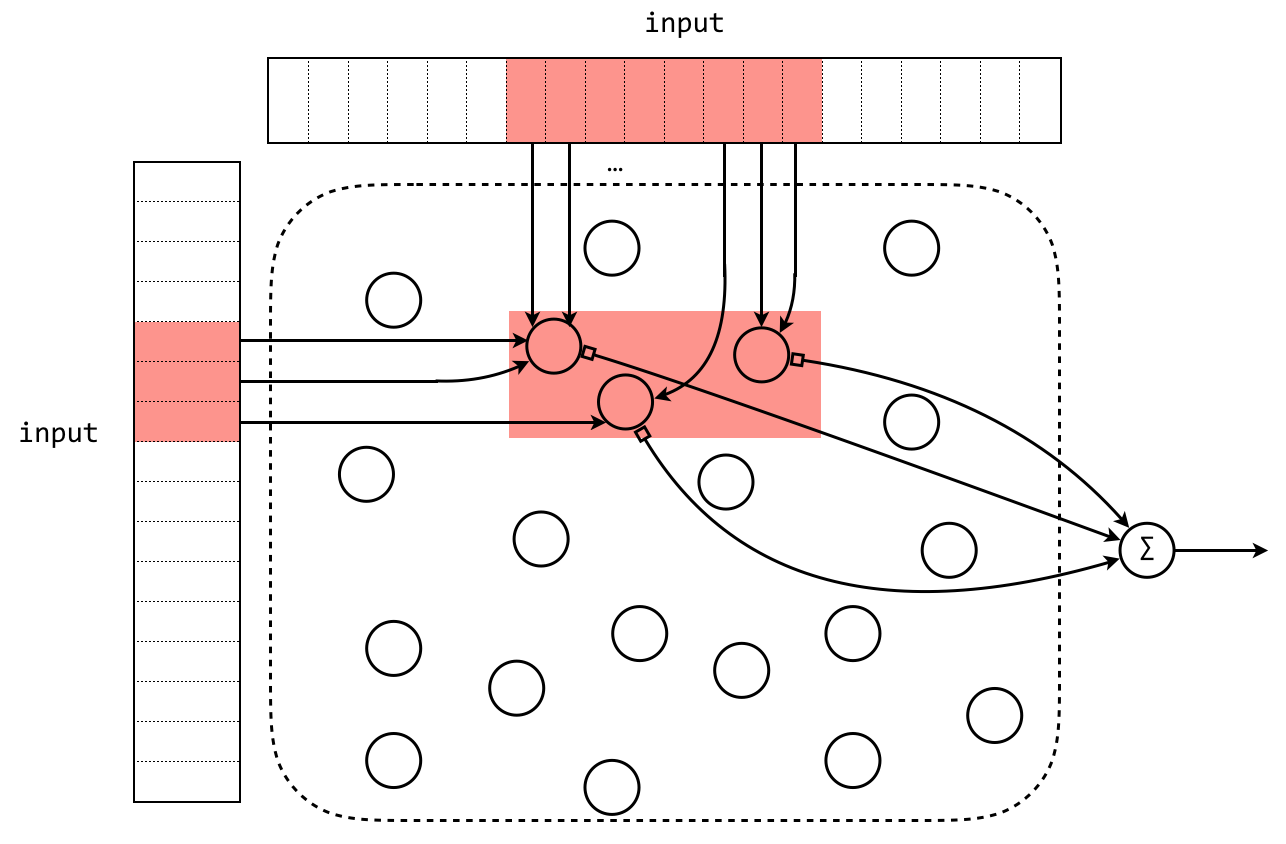}
\includegraphics[scale=0.4]{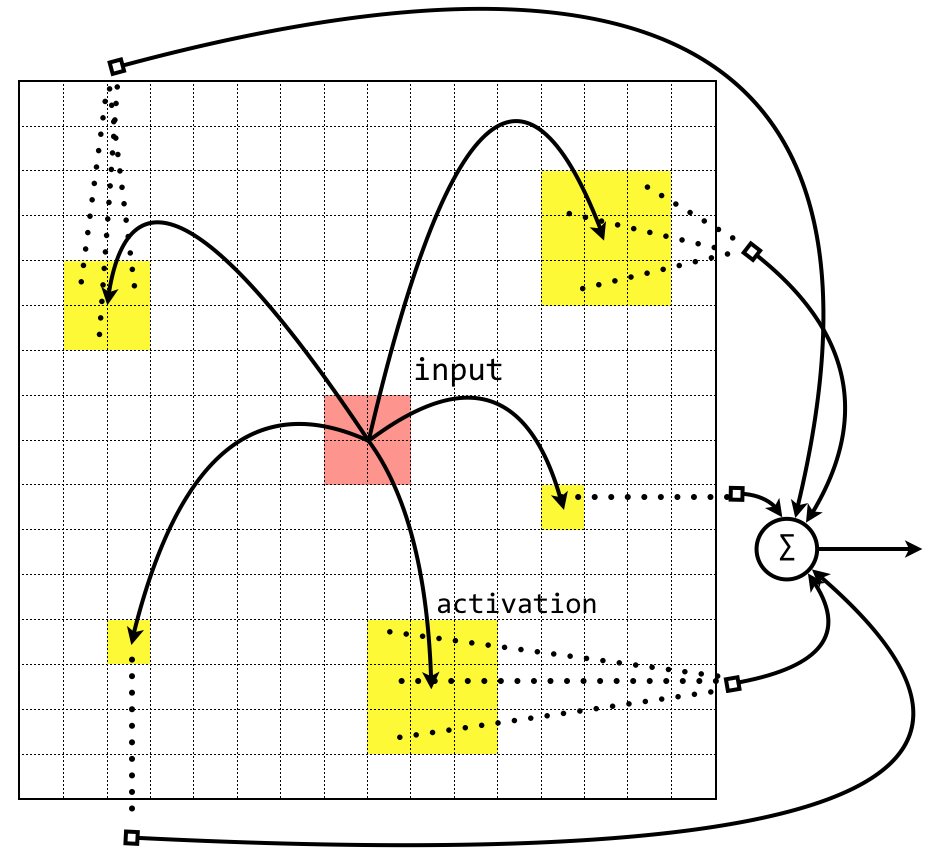}
\caption{A conceptual level comparison of two different perspectives}
\label{cmp}
\end{figure}

When it comes to implementation, the connectivism perspective to recognize CMAC as a neural network and the table referring perspective are equivalent. Fig. \ref{cmp} illustrates the difference at a conceptual level \cite{smith98}. To the upper part is a two input one output neural network structure, to the lower part is a two input one output table look-up structure.

An intuitive observation from both Fig. \ref{newcmac} and Fig. \ref{cmp} is that the number of weights will increase exponentially with the number of input variables. This problem brings out two challenges: 1) The storage of weights become space consuming; 2) The training process becomes difficult to converge and waiting time before termination will lengthen.

A previously used technique to solve the first challenge is called tile coding, which latter was developed to an adaptive version as well \cite{whiteson07}. The advantage of tile coding is that we can strictly control the number of features through tile split. Another commonly employed trick is called hashing. This technique is applied to CMAC in the 1990s, and maps a large virtual space $\mathit{A}$ to a smaller physical address $\mathit{A'}$. Many common hashing function $f_h$ can be used, for instance from MD5 or DES, which are fundamental methods in cryptography. However, how to reduce the collision for specific problems according to the data property is still considered an art. 

Literature \cite{smith98} introduced a hardware implementation which uses selected binary bits of indexes as the hashing code. Whereas other research, e.~g. \cite{wang96}, claims that due to the learning rate diminishing and slower convergence, hashing is not effective for enhancing the capability of approximation by CMAC. Therefore, many other attempts have been made, such as neighbor sequential, CMAC with general basis functions, and adaptive coding \cite{duan99}, which is a similar idea to hashing in the sense of weight space compression. 

\vspace{1cm}
\subsection{Modified Architectures}
Since simply increasing the CMAC size gives diminishing returns, two directions of modification are undertaken to push forward the research on CMAC. The \emph{first} consideration is to combine multiple low dimensional CMACs. The \emph{second} consideration is to introduce other properties, for example spline neurons, fuzzy coding or cooperative PID controller.

\begin{figure}[thpb]
\centering
\includegraphics[scale=0.45]{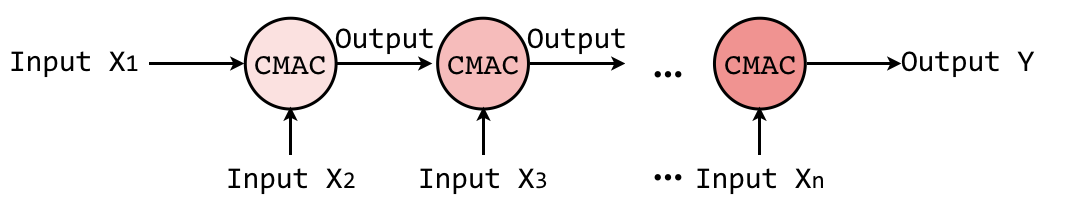}
\caption{Cascade CMAC architecture}
\label{ccmac}
\end{figure}

Cascade CMAC architecture was firstly proposed in 1997 for the purpose of printer calibration \cite{huang97} (Fig. \ref{ccmac}). Input variables are sequentially added to keep each of the CMAC component two dimensional.

Another method to combine multiple CMACs can be realized by voting technique. If we regard the Cascade CMAC model as a fusion of input information at a feature level, then the voting CMACs can be reckoned as a fusion at decision level. Each small CMAC just accept a subset of the whole input space. In this case an important antecedent is that input data is well partitioned. The reason to this requirement is that voting lift can only be achieved by heterogeneous expert  networks. Taken this into account, some prior knowledge of input data or unsupervised clustering techniques can be applied in this stage.

\begin{figure}[thpb]
\centering
\includegraphics[scale=0.4]{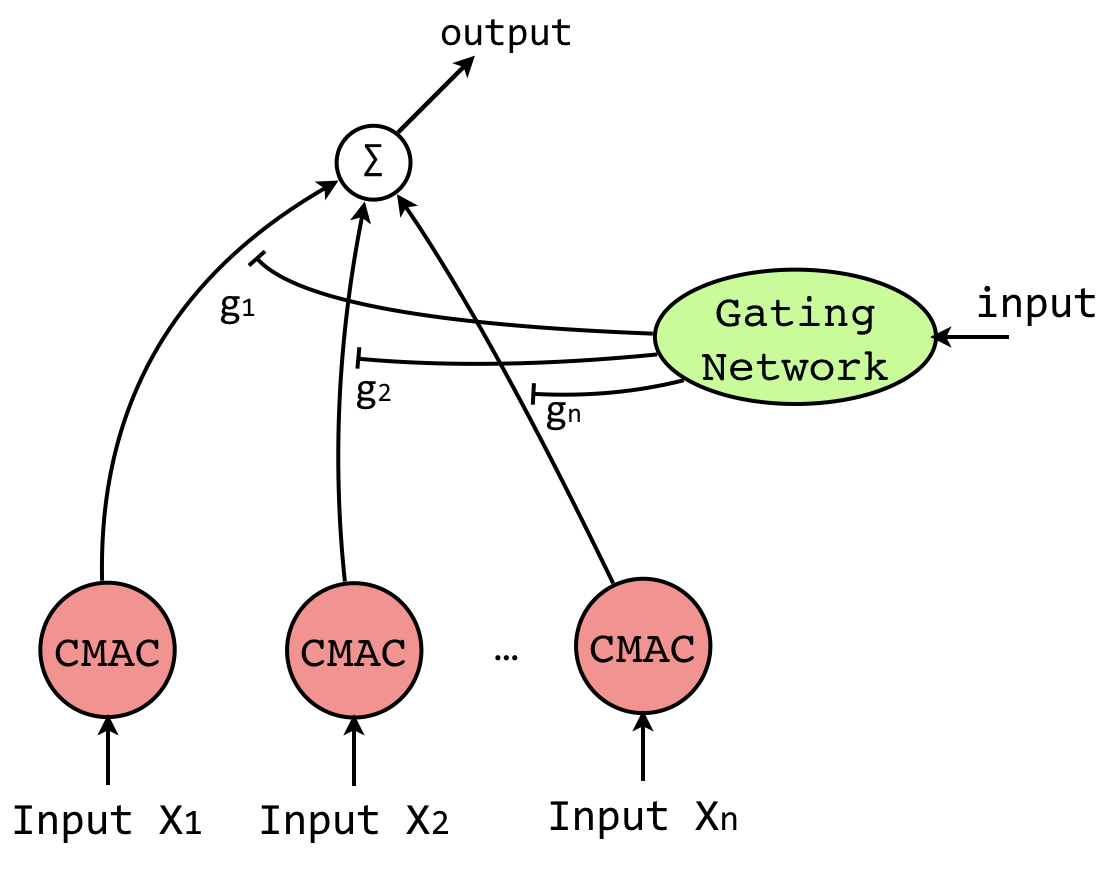}
\caption{Hierarchical CMAC architecture, adapted from \cite{tham}}
\label{hcmac}
\end{figure}

If we make more efforts for dimension reduction, multiple levels of voting can be used. Then the architecture can be reckoned as a Hierarchical CMAC (H-CMAC), which is described by Tham \cite{tham} in 1996. H-CMAC has several advantages, such as less storage space and fast adaptation for learning non-linear functions. In Fig. \ref{hcmac}, a two level H-CMAC is illustrated. It is noticeable that each conventional CMAC components in different layers plays different role. The gating network works at a higher level.

These architectures can be employed at the same time with more fundamental modifications for the second consideration. In 1992, Lane et al. \cite{lane92} descried high order CMAC as CMAC with binary neuron output replaced by spline functions. This modification brings about more parameters of splines, but makes the output derivable, which sometimes gives a better performance because the learning phase goes deeper. Sharing the idea to allow more meticulous transfer function, a similar modification can be made by introducing linguistic rules and fuzzy logic.

Linguistic CMAC (LCMAC) was proposed by He and Lawry based on label semantics, rather than mapping functions in 2009. A cascade of LCMAC series was further developed in 2015 \cite{he15}. Borrowing the terminology of ``focal element'' from evidence theory, the properties used to activate it is represented as membership function (usually trapezoidal) of several attributes. Therefore, for each input tuple, the excited neurons form a hypercube in the matrix of all the weights as memory. The responsive output can be distributionally depicted as:
$$
P(a|\mathbf{x}) = \prod\limits_{d=1}^N m_{x_d}(F_{d_i})
$$
where $P$ is the probability of some memory unit $a$ been activated given the input vector $\mathbf{x}$. $F$ denotes focal element. $d_i$ is the index of linguistic attributes. $m$ is the hidden weights for $F$ called ``mass assignment''.

Fuzzy CMAC (FCMAC) is yet another form of fuzzy coding. The intuition to use fuzzy representation is similar to using spline function. For most well defined problems, the nature of CMAC approximation is using multiple steps to emulate a smooth surface. Proper selection of fuzzy membership function would obviously relief the pressure of weight storage and training. From my understanding, FCMAC is an inverse structure of many established Neuro-Fuzzy Inference Systems. Usually, two extra fuzzy/defuzzy layers are added next to the association layer, the consequents can be Mamdani type, TSK type, weights, or a hybrid of them, e. g. Parametric-FCMAC (PFCMAC) \cite{moha09f}. More advanced FCMAC models, maybe inspired by spline methods, use interpolation to solve the discrete inference problem. In 2015, Zhou and Quek proposed FIE-FCMAC \cite{zhou15}, which adds fuzzy interpolation and extrapolation to a rigid CMAC structure. 

Recently, CMAC applications to more specific scenarios are studied. For example, for control of time-varying nonlinear system, a combination of Radial Basis Function network and the local learning feature of CMAC is proposed (RBFCMAC). It is reported that using RBFs can prevent parameter drift and accelerate synchronization speed to the changing system \cite{macnab16}. For this type of combination, beside RBF, Wavelet Neural Network (WNN), fuzzy rule neuron and recurrent mechanism \cite{lin13} or a mingle of them can also be employed with CMAC model simultaneously. Previous works, such as \cite{xing17}, have provided evidence that these features are effective for modeling complicated dynamic systems.

\begin{figure}[thpb]
\centering
\includegraphics[scale=0.5]{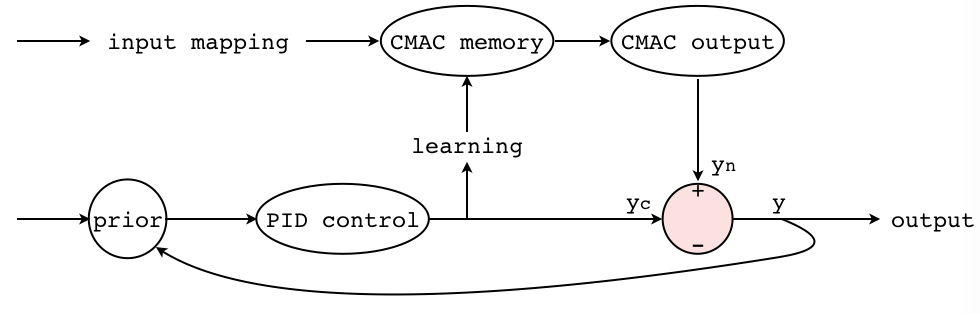}
\caption{A combined control system of CMAC and PID}
\label{combine}
\end{figure}

In a broader scenario, CMAC can be applied with other control systems as well (Fig. \ref{combine}). Traditionally, the role of CMAC at the primary stage is to assist the output of a main controller. As the training proceed, error between CMAC and the actual model decays. CMAC takes charge of the main controller. Back-forward signal acceptor or conjugated CMACs are often used to accelerate this process \cite{lin01}. More precisely, this arrangement is a change of information flow but not a change of architecture.

Similarly, CMAC structures that modifies the storage optimizing methods, for example quantization \cite{lu06} and multi-resolution \cite{menozzi97}, will only result in architectural difference from a hardware implementation sense. According to my personal understanding, they are the same thing regarding the conceptual structure, though these techniques are of sufficient interests to be discussed in Section \ref{lrn}.

\section{LEARNING ALGORITHM} \label{lrn}
The original form of learning proposed Albus is based on backward propagation of errors. The fast convergence of this algorithm is proved mathematically by succeeding researches. Specifically, the convergence rate is only governed by the size of receptive fields for association cells \cite{wong93}. Some study \cite{smith98} suggests that it would be useful to distinguish between target training and error training, despite they share the same mathematical form. In the weights updating rule,
\[
w_i(k+1)= w_i(k)+\alpha \frac{Error}{count(\mathit{A}^*)}\ x_i
\]
$k$ is the epoch times, $\alpha \in [0, 1]$ is the learning rate, $x_i$ is a state indicator of activation. The learning process is theoretically faster with a larger $\alpha$, while overshooting may occur. If the difference between output and the desired value $\Delta y = y-\hat{y}$ can well define the error, target training and error training will be equivalent. In certain cases, 
\[
Error= \frac{1}{2c}\ (y-\hat{y})^2
\]
is a popular cost function as well.

Based on the original learning algorithm, many developments are further derived. Most of them can be categorized into two directions of  improvement. The first relies on extra supervisory signals or value assignment mechanism based on statistics. The second endeavors to optimize the use of memory. From a more practical sense, the dichotomy can be understood as learning to adjust value of weights, and learning to adjust number or size of weights.


\subsection{Adjusting value of weights}
Facing the trade off between speed of convergence and learning stability, it is intuitive to consider using a relatively large $\alpha$ at the beginning, and slow down the weight adjustment near the optimum point. This improvement is called adaptive learning rate \cite{lin04}. It can be achieved by using two CMAC components, one as main controller, another as supervisory controller or compensated controller. Another way to achieve this adaption is by imposing a resistant term to the weight updating rule:
\[
w_i(k+1)= w_i(k)+\frac{\alpha_0}{c \alpha_i} \frac{Error}{count(\mathit{A}^*)}\ x_i
\]
In the above rule, $\alpha_0$ and $c$ are constants, $\alpha_i$ denotes the average activation times of the memory units that have been activated by training sample $i$.

For both fixed learning rate and adaptive learning rate, weights adjustment start from a randomized set of parameters. Experiments suggest that for sparse data, even if the training samples are within the local generalization range, perfect linear interpolation may not be achieved \cite{smith98}. As a result, the approximation may appears to have many small zigzag patterns. Therefore, the weight smoothing technique is proposed. After each iteration, the weights are globally adjusted according to:
\[
w_i(A^*_j)= (1-\delta)\ w_i(A^*_j)+\frac{\delta}{count(\mathit{A}^*)} \sum\limits^{count(\mathit{A}^*)}_{k=1} w_i(A^*_k)
\]
where $\delta$ is the proportional coefficient measuring the share of the weight with index $j$ that needs to be replaced by the average of all activated weights $\mathit{A}^*$.\\

While the weight smoothing technique tries to affect as much memory as possible in one iteration, repeated activation of the same units may not be a good thing. In 2003, research \cite{su03} pointed out that equally assign the error to each weight is not a meticulous method. During the learning process, if an associative memory is activated many times, which means many relevant samples are already learned, the weight should be more close to the desired value. In other words, the weight value is more ``credible". In this situation, less error should be assigned to it so that other memory units can learn faster. This rule is called learning based on credit assignment:
\[
w_i(k+1)= w_i(k)+ \frac{\alpha f(i)^{-1}}{\sum\limits^g_{i=1} f(i)^{-1}}\ (y-\sum\limits^g_{i=1} w_i(k))\ x_i
\]
where $g$ is a parameter regarding the degree of local generalization, $f(i)$ records the times memory unit $i$ has been activated. Further research has proved that the convergence is guaranteed with learning rate $\alpha<2$.

The hardware implementation of CMAC memory enables other interpretations of the associative architecture. If we recognize it as a self-organizing feature map (SOFM), competitive learning algorithm can be realized. With input variables $\bm{x}=\{x_1,\ x_2,\ ...,\ x_n\}$, the output can directly be the weight of the winning neuron. The weights updating rule can be formalized according to Hebbian theory:
\[
w_{\bm{x}}(k+1)= w_{\bm{x}}(k)+ \alpha\ (y(k)-w_{\bm{x}}(k))
\]
Note that the index of $\bm{x}$ can be time dependent. 

A modified version of the aforementioned rule involves not only inputs, but also errors as feedback:
\[
w_{\bm{x},\bm{e}}(k+1)= w_{\bm{x},\bm{e}}(k)+ \alpha\ (y(k)-w_{\bm{x},\bm{e}}(k))
\]
Using this learning mechanism, the SOFM-like CMAC is named MCMAC. In 2000, Ang and Quek \cite{ang00} proposed learning with momentum, neighborhood, and averaged fuzzy output for both CMAC and MCMAC. For CMAC and MCMAC with momentum, the weights updating rules can be written as:
\begin{numcases}{}
\Delta w_{j}(k+1) = \alpha\ \Delta w_{j}(k) + \eta\ \delta_j(k) y_j(k) \nonumber \\
\Delta w_{\bm{x},\bm{e}}(k+1)= \alpha\ \Delta w_{\bm{x},\bm{e}}(k)+\lambda(1-\alpha)[y(k)-w_{\bm{x},\bm{e}}(k)]\nonumber
\end{numcases}
where the first term represents a momentum, the second term is a back propagation term with learning rate $\eta$ and local gradient $\delta_j$. $j$ is the index for activated weights.

Therefore, given a sequential learning process, the aggregational weights adjustment can be derived from the above rules:
\begin{equation}
\begin{aligned}
\Delta w_{j}(k+1) &= \eta\ \sum\limits^k_{i=1} \alpha^{k-1} \delta_j(i) y_j(i)\\
&= -\eta\ \sum\limits^k_{i=1} \alpha^{k-1}\frac{\partial Error}{\partial w_{j}(k)} \nonumber
\end{aligned}
\end{equation}
When the sign of $\frac{\partial Error}{\partial w_{j}}$ keeps unchanged, $\Delta w_{j}$ is accelerating; while if the sign is reversing, $\Delta w_{j}$ will slow down to stabilize the learning process.

The neighborhood learning rule proposed by Ang and Quek \cite{ang00} serves the same purpose as weight smoothing technique. However, they used the Gaussian function to put more attention to those neurons surrounding the winning neuron instead of evenly adjusting each weight regardless of the distance. Weights updating rule for MCMAC with momentum and neighborhood in singular form is:
\[
\Delta w_{i}(k+1)= \alpha\ \Delta w_{i}(k)+h_{ij} [\lambda(1-\alpha)(y(k)-w_{i}(k))]
\]
where \[h_{ij} = \exp\Big(-\frac{|r_j-r_i|^2}{2\sigma^2}\ \Big)\] is the distance metric between neuron $j$ and the winning neuron $i$.

Beside using additional terms such as momentum and neighborhood, kernel method can also be applied to CMAC learning. In 2007, Kernel CMAC (KCMAC) was proposed \cite{hor,lane92} to reduce the CMAC dimension which usually hazard the training speed and convergence. KCMAC treats the association memory as the feature space of a kernel machine, thus weights can be determined by solving an optimization problem. The supervised learning using error $\bm{e}$ as slack variables is to achieve:
\begin{equation}
\begin{aligned}
&\min_{\bm{w}, \bm{e}}\ \frac{1}{2} \bm{w}^T \bm{w}+\frac{\beta}{2}\sum\limits_{i=1}^n \bm{e}_i^2\\
&s.t.\ \bm{w}^T \phi(u_i) + \frac{\beta}{2} \bm{e}_i \geqslant 1,\ \ i=1, 2, ..., n\nonumber
\end{aligned}
\end{equation}
where $\bm{w}$ denotes weights, coefficient $\beta$ serves as penalty parameter, $\phi(\cdot)$ is the mapping function and $K(\bm{u},\bm{w})=\phi(\bm{u})\cdot\phi(\bm{w})$ is the kernel function. 

The standard procedure to this problem is to solve the maxima-minima of Lagrangian function. Though other learning method can be employed as well, for instance, using Bayesian Ying-Yang (BYY) learning as proposed in 2013 by Tian et al \cite{tian13}. The key idea behind BYY learning can be represented by harmonizing the joint probability with different product form of Bayesian components. In this specific KCMAC case, $\bm{u}$ and output $\bm{z}$ are observable, while $\bm{w}$ is a hidden variable. The joint distribution can either be written as $ying$ form or $yang$ form.
\begin{numcases}{}
p_{ying}(\bm{u},\bm{z},\bm{w})=p(\bm{w}) p(\bm{u}|\bm{w})p(\bm{z}|\bm{u}, \bm{w}) \nonumber \\
p_{yang}(\bm{u},\bm{z},\bm{w})=p(\bm{u}) p(\bm{z}|\bm{u})p(\bm{w}|\bm{u}, \bm{z}) \nonumber\nonumber
\end{numcases}
Our goal is to maximize $\mathscr{H}(p_{ying}, p_{yang})$,
\[
\mathscr{H}(p_{ying}, p_{yang})= \iiint p_{ying}\ln p_{yang} \ d\bm{u}\,d\bm{z}\,d\bm{w}
\]
In practice, $\Delta \mathscr{H}$ is frequently calculated depending on how the conditional probabilities are estimated and maximum of $\mathscr{H}$ is achieved by searching heuristically.

\subsection{Adjusting number of weights}
Adjusting number of weights is chiefly realized by introducing multi-resolution and dynamic quantization techniques. As Section \ref{intro} has explained, CMAC was firstly used for real time control system. Consequently, the structural design fits the hardware implementation well. Many CMAC variants, such as LCMAC and FCMAC, inherit the memory units division with a lattice-based manner. Inputs are generally built on grids with equal space. This characteristic adds local constraints to the value of adjacent units.

If we consider this problem from a function approximation perspective, it is also rather intuitive that local complex shape needs a larger number of low-order elements to approach. Naturally, multi-resolution lattice techniques \cite{menozzi97} were proposed in the 1990s.

With our prior knowledge, some metrics can be used to determine the resolution, for example, the variation of output in certain memory unit areas, which can be formalized as following:
\[
resolution\ \propto \ v=\frac{1}{N} \sum\limits_{j=1}^N |y_j - \bar{y}|
\]
where $N$ is the number of output samples, $v$ is the variance. Other attempts use a tree structure to manage the resolution hierarchically. New high resolution lattice is generated only if the threshold of variance is exceeded.  

The concept of quantization is almost identical to resolution. The nuance may be that the terminology \emph{quantization} puts more emphasis on the discretion of continuous signal. Furthermore, increasing resolution will also cause the number of weights to grow. Quantization deals with a given memory capacity problem.

To the best of my knowledge, the idea of adaptive quantization was initially proposed in 2006 \cite{lu06}. The algorithm used to interpolate points looks on change of slopes, which is also similar to the variance metric discussed above. The input space is initialized with uniform quantization. For each point $x_i$, the slopes are calculated by neighbor points.
\[
\hat{slope}_j = \frac{f(x_j)-f(x_i)}{x_j - x_i},\ \ j=1,2,3...
\]
Here $f$ stands for the mapping function between input vector and association cells. The change of sign for corresponding directions indicates finer local structure, or fluctuation in other words, thus a new critical point is added to split the interval. The termination condition is that, within all intervals, the difference of output values should not exceed an experimental constant $\mathit{C}$ \cite{teddy07}. The adaptive quantization technique is later developed to the Pseudo Self Evolving CMAC (PSECMAC) model \cite{teddy08a}, which further introduced neighborhood activation mechanism.

\section{APPLICATION} \label{app}
Though CMAC was firstly proposed for manipulator control, during the past decades it has been proved effective in robotic arm control, fractionator control, piezoelectric actuator control \cite{peng03}, mechanic system, signal processing, intelligent driver warning system, fuel injection, canal control automation, printer color reproduction, ATM cash demand \cite{teddy11}, and many other fields \cite{duan99}. This can be attributed to the fast learning and stable performance of CMAC models. Moreover, engineering-oriented software and toolkit also help to promote the application of CAMC. In the following part, CMAC application to two emerging engineering fields are elaborated.

\subsection{Financial Engineering}
The practice side of financial engineering has employed various instruments to model the price and risk of bond, stock, option, future and other derivatives. In most cases, historic data can be partitioned into chunks of selected time span to enable a supervised learning process. In 2008, Teddy et al \cite{teddy08} proposed an improved CMAC variant (PSECMAC) to model the price of currency exchange American call option. Three variables are taken into consideration: difference between strike price and current price, time of maturity, and pricing volatility. Thus the pricing function is formulated as:
\[
C_0 = f(S_0-X, T, \sigma_{30})
\]

The article reported PSECMAC as the best-performing model among several CMAC-like systems. It is also reported that most of the CMAC models produce better result than using Black-Scholes model in sense of RMSE. Though for American option, Black-Scholes model is not a good benchmark because it is sensitive to details of calculation.

This work further constructed an arbitrage system based on the pricing model. Positions are adjusted according to the Delta hedging ratio. Experiments suggested the model has a marginal positive ROI, omitting transaction costs.

\subsection{Adaptive Control}
The application of CMAC on commercial devices may be more advantageous. The pressure to reduce cost and demand of embedded controller make CMAC-like models a good choice. Recently, studies have been carried on adaptive control of disabled wheelchair and seatless unicycle \cite{lyy}. This type of problems can be formalized as controlling a set of variables in certain range (e. g. speed and balance angle) with a set of unknown and varying variables (e. g. friction to the ground and weight of the rider).

Li et al \cite{lyy} proposed a TSK type FCMAC to synthesis the equations of adaptive steering control. The back-stepping error is associated with the torque $\tau$, which can simultaneously effect on critical moments.
\begin{numcases}{}
\ddot{\theta} = \bar{A}(\theta,\dot{\theta}) + \bar{B}(\theta)(\tau-\mu\dot{\phi}-c\ \texttt{sgn}(\dot{\phi})) \nonumber \\
\ddot{\phi} = \bar{C}(\theta,\dot{\theta}) + \bar{D}(\theta)(\tau-\mu\dot{\phi}-c\ \texttt{sgn}(\dot{\phi}))\nonumber
\end{numcases}
As the output adapting to the moment parameters, the balance angle is controlled near zero.

The performance of FCMAC is benchmarked with a linear-quadratic regulator for differential equations to describe the state of the motor. Simulations suggest that LQR is not able to converge speed and angle of balance, while FCMAC provides satisfactory result.
\vspace{0.3cm}

\section{DISCUSSIONS} \label{disc}
Referring to Section \ref{app}, CMAC was proved to be effective in many classic control problem and has been applied to emerging engineering problems. However, this model seems have encountered a bottleneck because of the lack of fundamental breakthrough during the past decade. Nowadays, issues discussed are mainly focused on trivial modification on memory structure and learning algorithm. The framework of error propagation or minimization of loss function is kept unchanged.

According to my understanding, the limitation of current CMAC models can be ascribe to its over simplification of the cerebellum structure. Therefore, the next generation cerebellar model may adopt new discoveries from neuroscience. For example, the associative memory cells may take different roles rather than been treated identically. In fact, anatomical models usually feature several types of elementary cerebellar processing units. Meanwhile, the theory of Spike Timing Dependent Plasticity (STDP) suggests that the learning process of firing neurons may be ordered \cite{cap}. This feature can introduce far more complexity to the current learning algorithm.

\end{document}